\documentclass[11pt]{article}

\usepackage[T1]{fontenc}
\usepackage[utf8]{inputenc}
\usepackage{lmodern}

\usepackage{amsmath,amssymb}
\usepackage{graphicx}
\usepackage{booktabs}
\usepackage{microtype}
\usepackage[hidelinks]{hyperref}
\usepackage[margin=1in]{geometry}

\usepackage{caption}

\title{Input-Adaptive Visual Preprocessing for Efficient Fast Vision-Language Model Inference}

\author{
Putu Indah Githa Cahyani\textsuperscript{1},
Komang David Dananjaya Suartana\textsuperscript{1},
Novanto Yudistira\textsuperscript{1}\\
\textsuperscript{1}Faculty of Computer Science, University of Brawijaya, Malang, Indonesia\\
\texttt{indahgitha@student.ub.ac.id},
\texttt{kmdavidds@student.ub.ac.id},
\texttt{yudistira@ub.ac.id}
}

\date{} 

\begin{document}
\maketitle

\begin{abstract}
Vision-Language Models (VLMs) have demonstrated strong performance on multimodal reasoning tasks, but their deployment remains challenging due to high inference latency and computational cost, particularly when processing high-resolution visual inputs. While recent architectures such as FastVLM improve efficiency through optimized vision encoders, existing pipelines still rely on static visual preprocessing, leading to redundant computation for visually simple inputs. In this work, we propose an adaptive visual preprocessing method that dynamically adjusts input resolution and spatial coverage based on image content characteristics. The proposed approach combines content-aware image analysis, adaptive resolution selection, and content-aware cropping to reduce visual redundancy prior to vision encoding. Importantly, the method is integrated with FastVLM without modifying its architecture or requiring retraining. We evaluate the proposed method on a subset of the DocVQA dataset in an inference-only setting, focusing on efficiency-oriented metrics. Experimental results show that adaptive preprocessing reduces per-image inference time by over 50\%, lowers mean full generation time, and achieves a consistent reduction of more than 55\% in visual token count compared to the baseline pipeline. These findings demonstrate that input-aware preprocessing is an effective and lightweight strategy for improving deployment-oriented efficiency of vision-language models. To facilitate reproducibility, our implementation is provided as a fork of the FastVLM repository, incorporating the files for the proposed method, and is available at https://github.com/kmdavidds/mlfastlm.
\end{abstract}


\section{Introduction}
Recent advances in Large Language Models (LLMs) have significantly
transformed artificial intelligence by enabling strong reasoning,
contextual understanding, and generalization across a wide range of
natural language tasks.This progress is largely driven by the scaling of
model parameters and training data, which leads to emergent
capabilities, as demonstrated by early milestones such as GPT-3 {[}1{]}.
Subsequent works refined model architectures and training strategies to
improve parameter efficiency while maintaining competitive performance
{[}2{]}. In parallel, prompting and instruction tuning enhanced
adaptability, allowing LLMs to generalize to unseen tasks with minimal
supervision {[}3{]}. However, despite these advances, LLMs remain
limited to textual inputs, constraining their applicability in tasks
that require visual understanding.

To overcome the perceptual limitations of text-only models, recent
research has shifted toward Vision-Language Models (VLMs) that jointly
reason over visual and textual modalities. A foundational contribution
is contrastive vision-language pretraining, exemplified by CLIP, which
aligns image and text representations using large-scale paired data and
enables strong zero-shot transfer across tasks {[}4{]}. This paradigm
has become a cornerstone of multimodal representation learning. Building
on it, recent VLMs integrate pretrained LLMs to inherit reasoning and
generation capabilities. For instance, Flamingo supports few-shot
multimodal learning through interleaved visual and textual conditioning
{[}5{]}, while BLIP-2 reduces training cost through lightweight bridging
modules {[}3{]}. Together, these models position VLMs as a scalable
extension of LLMs toward general-purpose multimodal intelligence.

While vision-language models exhibit strong representational power,
their practical deployment remains challenging due to computational and
memory overhead. Contemporary VLMs combine high-capacity visual encoders
with large language backbones, leading to increased inference latency
and energy consumption, particularly for high-resolution visual inputs
{[}3{]}, {[}5{]}. These constraints hinder adoption in real-time and
resource-constrained settings such as edge devices. Empirical evidence
suggests that scaling model size alone does not yield proportional
efficiency gains and may exacerbate deployment bottlenecks {[}5{]},
{[}6{]}. Although parameter-efficient tuning reduces training cost,
inference-time inefficiencies caused by dense visual tokens remain
largely unaddressed {[}7{]}. As a result, improving deployment-oriented
VLM efficiency remains a critical research challenge.

Recent efforts to improve VLM efficiency have focused on
architecture-level optimizations that balance performance and
computational cost. A notable example is FastVLM, which introduces a
hybrid vision encoder to reduce visual token counts and encoding latency
for high-resolution inputs while preserving competitive benchmark
performance {[}8{]}. Unlike compression or pruning methods, FastVLM
achieves efficiency gains by redesigning the vision encoding pipeline.
Complementary approaches explore selective activation of visual experts
to adapt model capacity to input complexity {[}9{]}. While effective,
these methods primarily address model-level efficiency and overlook the
potential of input-aware preprocessing strategies for further reducing
inference cost. Fig. 1 illustrates the performance trade-offs of FastVLM
and representative efficient VLMs under varying visual token budgets.

\begin{figure}[htbp]
    \centering
    \includegraphics[width=\linewidth]{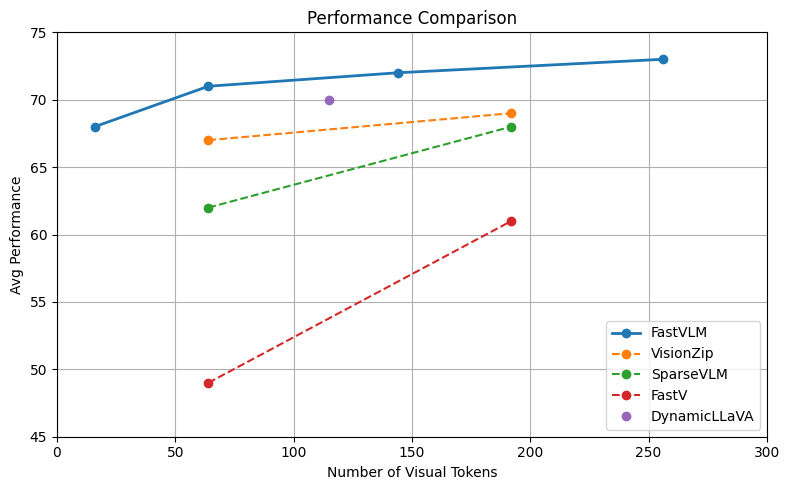}
    \caption{Performance comparison of FastVLM and representative efficient vision-language models across varying numbers of visual tokens.}
    \label{fig:image9}
\end{figure}

In this work, we address the limitations of static visual preprocessing
in existing vision-language pipelines by proposing an adaptive visual
preprocessing approach. Instead of applying uniform preprocessing to all
images, our method adjusts preprocessing operations based on input
characteristics, such as visual complexity and information density. This
strategy reduces redundant visual information while preserving features
critical for multimodal reasoning. The proposed approach is integrated
with FastVLM as an efficient backbone without modifying its core
architecture. We evaluate the method on standard vision-language
benchmarks and report both performance and efficiency metrics,
demonstrating its effectiveness in improving deployment-oriented VLM
efficiency. Rather than proposing a new architectural or learning-based
mechanism, this work deliberately investigates how far inference
efficiency can be improved through lightweight, input-level adaptation
alone.

This work explicitly focuses on improving inference efficiency of
vision-language models rather than enhancing task-level accuracy or
benchmark performance. The proposed adaptive preprocessing is designed
to reduce computational overhead, visual token redundancy, and inference
latency during deployment. Consequently, this study does not aim to
improve visual question answering accuracy, reasoning capability, or
language generation quality beyond the baseline FastVLM model. Instead,
it investigates how far deployment-oriented efficiency can be improved
through lightweight, input-level adaptation alone, under the assumption
that task performance should remain largely stable.

\section{Related Works}
Recent progress in vision-language modeling has been strongly driven by
instruction tuning, which aligns pretrained language models with visual
inputs for multimodal reasoning. Frameworks such as LLaVA demonstrate
that large-scale multimodal instruction data can significantly improve
visual question answering and conversational abilities without modifying
the language backbone {[}10{]}. Subsequent studies show that these gains
depend not only on model size but also on data composition and prompt
design {[}11{]}. InstructBLIP further unifies heterogeneous
vision-language tasks under an instruction-driven framework, improving
generalization across diverse visual and linguistic conditions {[}12{]}.
Together, these works establish instruction tuning as a key paradigm for
multimodal alignment.

Despite improved alignment, deployment of vision-language models remains
limited by the high cost of visual encoding, especially for
high-resolution inputs. Lightweight architectures and mobile-oriented
models reduce latency and memory usage under constrained settings
{[}13{]}, while encoder redesigns such as ConvLLaVA employ hierarchical
backbones to reduce redundant visual tokens without sacrificing spatial
structure {[}14{]}. Scalable encoders like ViTamin further highlight
that stronger visual representations can partially alleviate downstream
token inefficiencies {[}15{]}.

Recent efficiency-focused research directly targets visual token
redundancy. Token pruning and merging methods dynamically remove less
informative tokens, substantially reducing attention complexity with
minimal accuracy loss {[}16{]}, {[}17{]}. Sparsification approaches
further adapt token budgets based on relevance or saliency, showing that
dense visual representations are often unnecessary for effective
multimodal reasoning {[}18{]}. To address high-resolution inputs,
methods such as LLaVA-UHD and dynamic context sparsification reorganize
or prune vision-language context to balance perceptual fidelity and
efficiency, though they typically require architectural or
inference-level modifications {[}19{]}, {[}20{]}.

In addition to computational cost, several studies emphasize that
inference-time efficiency is increasingly critical for real-world
deployment scenarios, such as interactive systems and edge devices.
These works highlight that latency, memory footprint, and energy
consumption often dominate practical constraints, motivating efficiency
improvements that can be applied without extensive retraining or
architectural redesign {[}13{]}, {[}17{]}, {[}20{]}.

In contrast, relatively little work explores efficiency gains at the
preprocessing stage. Most pipelines rely on static resizing regardless
of image content, leading to unnecessary encoding of visually redundant
regions. This gap motivates the proposed adaptive visual preprocessing
approach, which reduces redundancy prior to visual encoding by adjusting
preprocessing based on input characteristics. Integrated with FastVLM,
this strategy complements existing model-level optimizations and
improves deployment-oriented efficiency without altering the core
architecture.

\section{Methods}

The proposed method is motivated by the limitations of existing
vision-language pipelines that rely on static visual preprocessing,
which leads to unnecessary computation and increased inference latency.
Although FastVLM significantly improves efficiency through an optimized
vision encoder, it still processes all image regions at a fixed
resolution, regardless of their visual relevance or complexity. This
results in redundant visual tokens, particularly for images containing
large background regions or low information density. To address these
challenges, we propose an adaptive visual preprocessing framework that
operates prior to vision encoding. The proposed method dynamically
adjusts image resolution and spatial coverage based on content
characteristics, enabling more efficient use of the FastVLM backbone.
Fig. 2 provides an overview of the proposed system.

\begin{figure}[htbp]
    \centering
    \includegraphics[width=\linewidth]{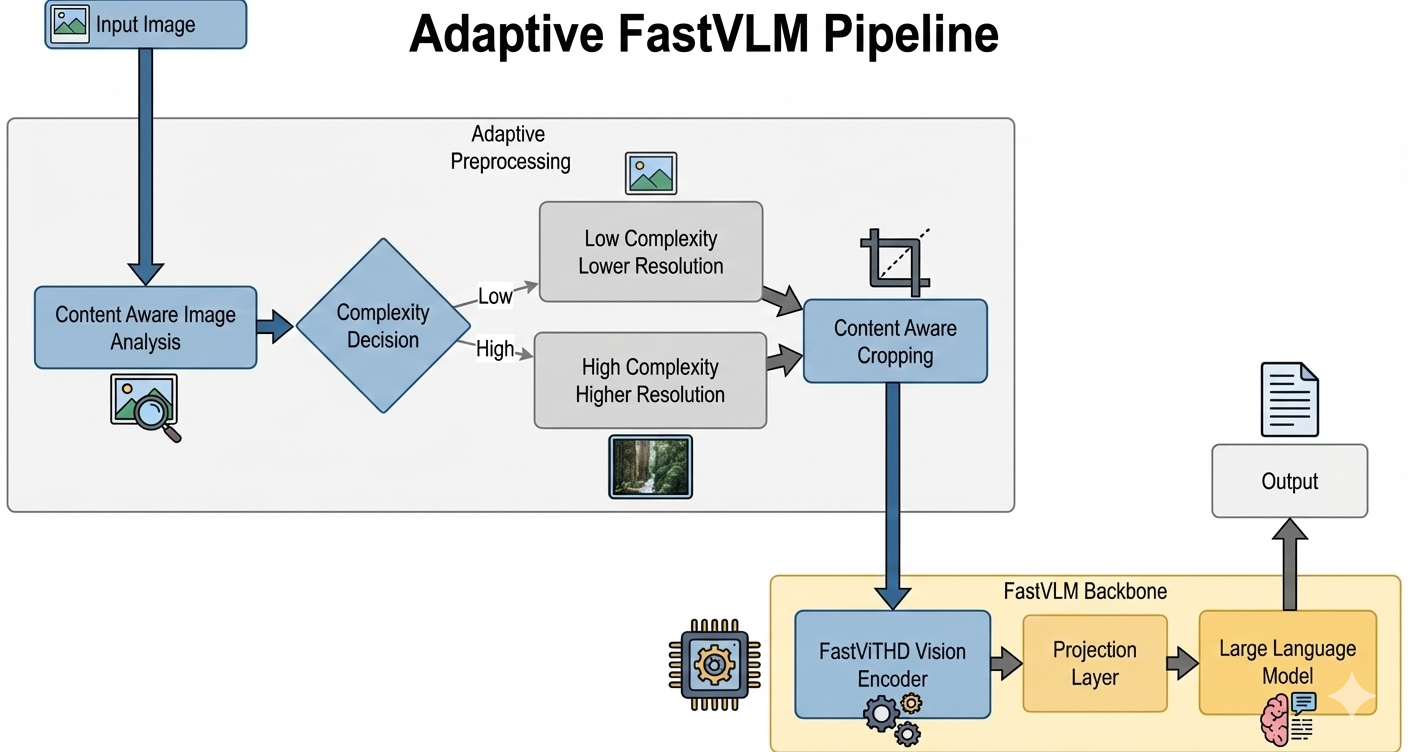}
    \caption{High-level overview of the proposed adaptive preprocessing
pipeline integrated with FastVLM.}
    \label{fig:image9}
\end{figure}

\emph{A. Content-Aware Image Analysis}

The first stage of the proposed pipeline performs content-aware image
analysis to estimate the visual complexity of the input image. Instead
of relying on computationally expensive models, this stage employs
lightweight image processing techniques to extract coarse indicators of
visual structure. Specifically, the input image is converted to
grayscale and analyzed using edge density or entropy-based measurements.

Edge density reflects the concentration of high-frequency visual
patterns, which commonly correspond to text regions, tables, or
fine-grained structures in document images. Similarly, entropy provides
a measure of pixel intensity variation, capturing the amount of visual
information present in the image. These measures serve as efficient
proxies for visual complexity, allowing the system to distinguish
between visually simple inputs, such as documents with sparse text or
large empty regions, and complex inputs containing dense textual or
tabular layouts. By using low-cost heuristics rather than learned
models, the image analysis stage introduces negligible overhead while
providing sufficient information to guide subsequent preprocessing
decisions. This design choice aligns with the deployment-oriented
objective of minimizing additional computation during inference.

The thresholds used to categorize images into low, medium, and high
complexity are empirically selected based on preliminary inspection of
edge density, entropy, and coarse text detection responses over document
images. In our implementation, complexity scores are normalized to the
range {[}0,1{]}, with images below a low-complexity threshold of 0.25
assigned to aggressive downscaling and content-aware cropping, and
images above a high-complexity threshold of 0.6 processed at higher
resolutions to preserve fine-grained visual details.

Lower threshold values favor more aggressive resolution reduction and
spatial cropping, leading to higher efficiency gains but a potentially
increased risk of visual information loss. Conversely, higher thresholds
prioritize visual fidelity at the cost of reduced efficiency
improvements. In practice, we observe that the efficiency--quality
trade-off evolves smoothly: moderate variations around the selected
threshold values lead to gradual changes in visual token count rather
than abrupt degradation. This behavior indicates that the proposed
method is not overly sensitive to precise threshold tuning, making it
robust and well suited for deployment scenarios where fixed, heuristic
parameter selection is preferred over learned or dynamically optimized
control mechanisms.

\emph{B. Adaptive Resolution Selection}

Based on the estimated image complexity, the adaptive resolution
selection component dynamically determines the appropriate input
resolution for vision encoding. Unlike conventional pipelines that apply
a fixed resolution to all images, the proposed method adjusts resolution
according to content characteristics.

For visually simple images, lower input resolutions are selected, as
high-resolution encoding would generate excessive visual tokens without
providing additional semantic benefit. Conversely, images identified as
high complexity are processed at higher resolutions to preserve
fine-grained visual details necessary for accurate document
understanding. This adaptive strategy directly impacts the number of
visual tokens generated by the vision encoder. Since visual token count
scales with input resolution, reducing resolution for simple images
leads to fewer tokens and lower computational cost during both vision
encoding and language model prefilling. At the same time, selective
high-resolution processing ensures that complex images retain sufficient
visual fidelity. As illustrated in Fig. 3, this resolution selection
mechanism enables the system to allocate computation selectively,
avoiding unnecessary processing while maintaining robustness for
challenging inputs.

\begin{figure}[htbp]
    \centering
\includegraphics[width=\linewidth]{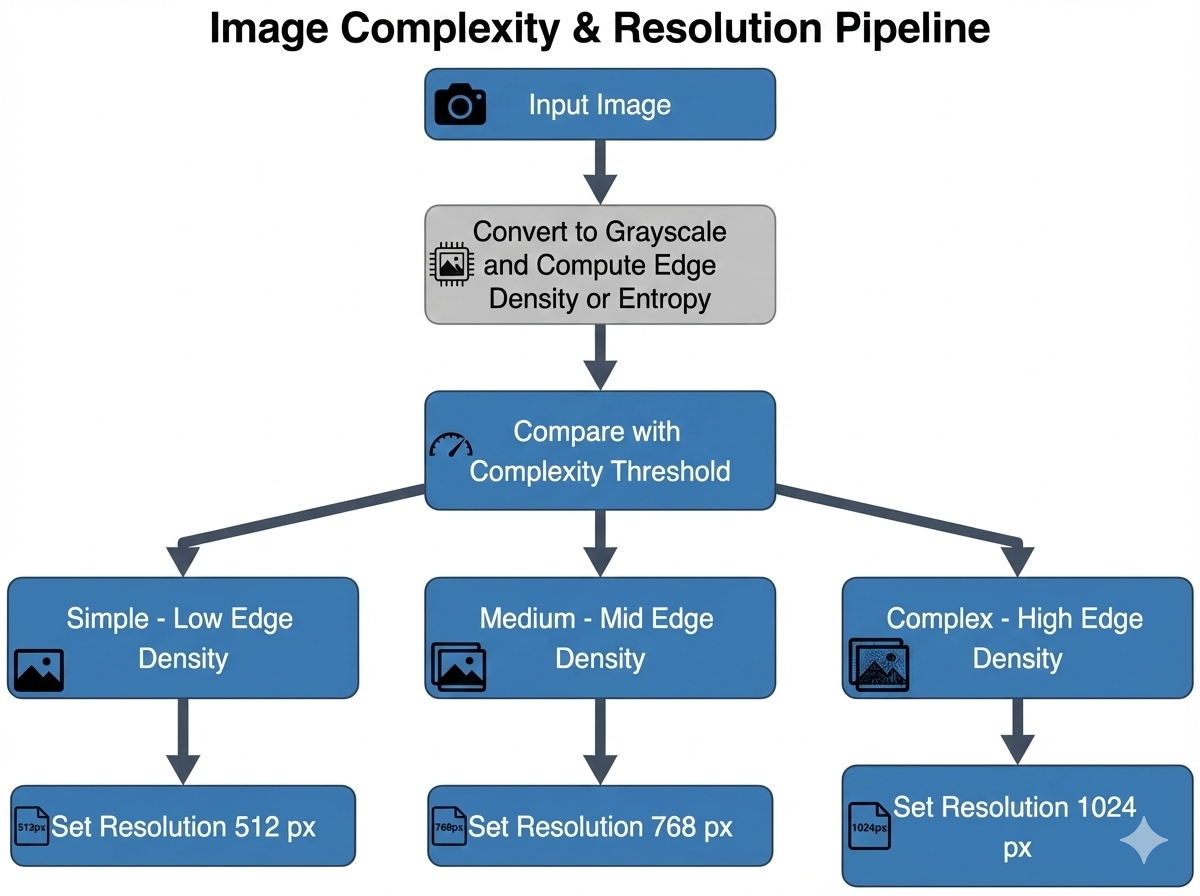}
    \caption{Adaptive resolution selection based on image complexity, where
edge density or entropy is used to categorize inputs into low, medium,
and high complexity levels, followed by dynamic resolution assignment.}
    \label{fig:image9}
\end{figure}

\emph{C. Content-Aware Cropping}

In addition to resolution adaptation, the proposed method applies
content-aware cropping to further reduce visual redundancy. Document
images often contain uninformative regions such as margins, backgrounds,
or empty spaces that do not contribute to vision-language reasoning.
Encoding these regions increases computational cost without improving
model performance.

To address this issue, the cropping component identifies spatial regions
that contain meaningful visual content. After grayscale conversion or
edge detection, pixels corresponding to low-information areas are
filtered out, and a bounding box enclosing content-dense regions is
computed. The image is then cropped to this bounding box, optionally
followed by resizing to match the selected resolution. By restricting
vision encoding to semantically relevant regions, content-aware cropping
reduces the spatial extent of the input while preserving important
visual structures. This targeted preprocessing further decreases visual
token generation and enables the vision encoder to focus computation on
regions most likely to influence downstream multimodal reasoning. The
cropping process is illustrated in Fig. 4.

\begin{figure}[htbp]
    \centering
\includegraphics[width=\linewidth]{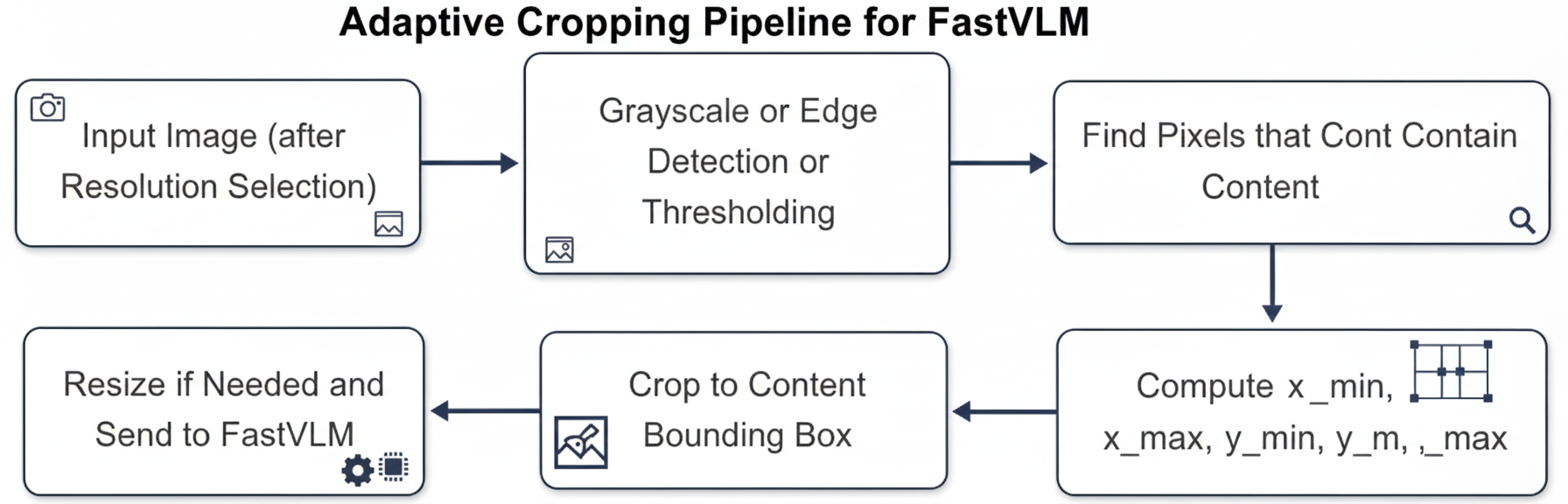}
    \caption{ Content-aware cropping pipeline that identifies informative
regions by detecting content pixels, computes the bounding box of
relevant areas, and crops the image before forwarding it to the FastVLM
backbone.}
    \label{fig:image9}
\end{figure}

\emph{D. Integration with FastVLM Backbone}

The proposed adaptive preprocessing pipeline is designed as a modular
front-end that integrates seamlessly with the FastVLM backbone. After
preprocessing, the resulting image is passed to the FastViTHD vision
encoder, followed by the projection layer and the large language model,
without any modification to model parameters or architecture. This
design ensures that efficiency gains are achieved solely through
input-level optimization. By decoupling preprocessing from model
internals, the proposed method remains compatible with existing FastVLM
deployments and can be readily applied to other vision-language models
that rely on token-based visual encoding.

Although this study integrates adaptive preprocessing with FastVLM, the
proposed method is not specific to FastVLM. The preprocessing operates
entirely at the input level and produces standard image tensors that can
be directly consumed by other vision-language models, such as LLaVA or
BLIP-2, which rely on token-based visual encoders. As long as the
downstream model processes visual inputs through fixed-resolution patch
embeddings or similar mechanisms, adaptive resolution selection and
content-aware cropping can be applied without architectural
modification.

\section{Experimental Setup}
\emph{A. Model Configuration}

All experiments are conducted using FastVLM as the vision-language
backbone. We evaluate FastVLM variants equipped with different language
model sizes, including Qwen2-0.5B and Qwen2-1.5B, to examine the
robustness of the proposed method across model scales. The vision
encoder is fixed to FastViTHD, which is specifically designed for
efficient high-resolution visual encoding. Importantly, the proposed
adaptive preprocessing method does not modify the FastVLM architecture;
all experiments use the same pretrained FastVLM checkpoints, ensuring a
fair comparison between static and adaptive preprocessing pipelines.

\emph{B. Dataset}

We evaluate the proposed method on a small, controlled subset of the
DocVQA dataset, a standard benchmark for document-based visual question
answering {[}21{]}. DocVQA consists of high-resolution, text-rich
document images that require fine-grained visual understanding and
accurate text localization, making it well suited for assessing adaptive
resolution selection and content-aware cropping strategies.

In this work, we use 32 document images paired with their corresponding
question-answer annotations. This subset is intentionally selected to
support a controlled, per-image efficiency evaluation, where each image
is processed using both the baseline FastVLM pipeline and the proposed
adaptive preprocessing method. Since the proposed method does not
involve learning or parameter estimation, the evaluation focuses on
controlled per-sample efficiency behavior rather than statistical
generalization, making a small but paired evaluation sufficient for the
scope of this study. While the evaluation is conducted on a limited
number of samples, this setup is adequate for examining efficiency
trends under identical conditions. Larger-scale evaluations are left for
future work.

\emph{C. Evaluation Protocol and Metrics}

We evaluate the proposed adaptive preprocessing method using a
controlled inference-time protocol. All experiments are conducted in an
inference-only setting, where the FastVLM model parameters remain fixed.
The proposed method is compared against the baseline FastVLM pipeline
using identical model checkpoints, prompts, and hardware configurations.
For each evaluated image, we perform a paired comparison, where the same
input is processed once using the baseline preprocessing and once using
the proposed adaptive preprocessing. This setup ensures that any
observed differences in efficiency metrics are solely attributed to the
preprocessing strategy, rather than model or input variations. The
evaluation focuses on the following efficiency-oriented metrics:
\

\begin{enumerate}
\def\labelenumi{\arabic{enumi}.}
\item
  \begin{quote}
  Inference Time. We measure the end-to-end inference time for each
  image, including vision encoding and language model generation. This
  metric reflects the overall latency experienced during deployment.
  \end{quote}
\item
  \begin{quote}
  Mean Full Generation Time. In addition to per-image latency, we report
  the average full generation time across all evaluated samples. This
  aggregate metric provides a stable summary of efficiency improvements
  introduced by the proposed method.
  \end{quote}
\item
  \begin{quote}
  Visual Token Reduction. We measure the relative reduction in the
  number of visual tokens generated by the vision encoder compared to
  the baseline. Since visual tokens directly impact both computation and
  memory cost, this metric captures the core efficiency benefit of
  adaptive resolution selection and content-aware cropping.
  \end{quote}
\item
  \begin{quote}
  Visual Quality Analysis. To verify that efficiency gains are not
  achieved at the expense of severe visual degradation, we additionally
  analyze the relationship between visual token count and visual quality
  scores after adaptive preprocessing. This analysis is used as a
  supporting validation rather than an optimization objective and aims
  to assess whether substantial token reduction preserves visually
  meaningful content.
  \end{quote}
\end{enumerate}

\section{Results}

\begin{quote}
This section presents a detailed evaluation of the proposed adaptive
preprocessing method compared to the baseline FastVLM pipeline. The
evaluation focuses on inference efficiency and computational cost,
measured through per-image inference time, mean full generation time,
and visual token reduction. All results are obtained under identical
inference settings using a paired evaluation protocol, where each image
is processed once with the baseline pipeline and once with the proposed
method.

\emph{A. Inference Time}

\begin{figure}[htbp]
    \centering
\includegraphics[width=\linewidth]{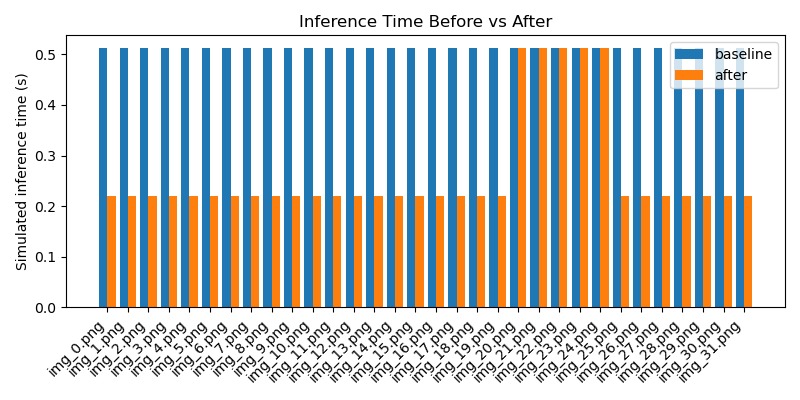}
    \caption{ Per-image inference time comparison between the baseline FastVLM
pipeline and the proposed adaptive preprocessing method.}
    \label{fig:image9}
\end{figure}

Fig. 5 compares the per-image inference time between the baseline and
the proposed method. Under static preprocessing, the baseline FastVLM
pipeline exhibits a narrow inference time range of approximately
0.50--0.52 seconds per image, indicating that similar computational cost
is incurred regardless of the visual complexity of the input. In
contrast, the proposed adaptive preprocessing method significantly
reduces inference time for most evaluated samples, with observed values
concentrated around 0.21-0.23 seconds per image. This corresponds to an
average absolute reduction of approximately 0.29 seconds per image,
representing a relative reduction of 55-60\% compared to the baseline.

Notably, a small subset of images exhibits inference times clThe
distribution of inference times reveals that efficiency gains are not
uniform across all inputs. A small subset of images exhibits inference
times closer to the baseline, which can be attributed to images
classified as high complexity during the content-aware analysis stage.
These images are intentionally processed at higher resolutions to
preserve fine-grained visual details, resulting in computational cost
comparable to the baseline. This behavior demonstrates that the proposed
method selectively reduces computation where appropriate, rather than
applying aggressive downscaling indiscriminately.

\emph{B. Mean Full Generation Time}

Fig. 6 reports the mean full generation time aggregated across all
evaluated samples. The baseline FastVLM pipeline achieves an average
full generation time of approximately 4.0 seconds, reflecting the
combined cost of vision encoding and language model decoding. When
adaptive preprocessing is applied, the mean full generation time is
reduced to approximately 3.8-3.9 seconds, resulting in an average
reduction of 0.1-0.2 seconds. While this reduction is smaller in
magnitude than the per-image inference time improvement, it remains
consistent across the evaluation set.

\begin{figure}[htbp]
    \centering
\includegraphics[width=\linewidth]{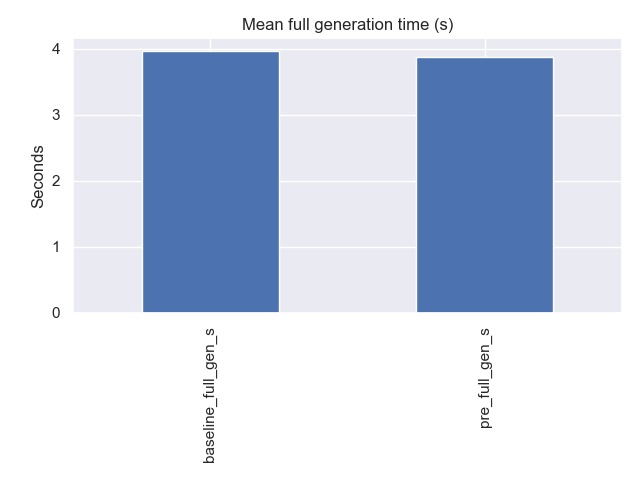}
    \caption{ Mean full generation time of the baseline and the proposed
adaptive preprocessing across the evaluated samples.}
    \label{fig:image9}
\end{figure}

This difference in relative improvement can be explained by the
structure of the FastVLM pipeline. Vision encoding constitutes only a
portion of the overall generation process, while the language model
decoding stage dominates the later steps. Consequently, reductions in
vision-side computation partially translate to end-to-end generation
time improvements, but do not scale linearly. Nevertheless, the observed
reduction in mean generation time confirms that lowering visual
computation cost contributes positively to overall system efficiency,
even when language model decoding remains a significant bottleneck.

\emph{C. Visual Token Reduction}

Fig. 7 shows the relative reduction in the number of visual tokens
generated by the vision encoder for each evaluated image. Compared to
the baseline pipeline, which processes all images at a fixed resolution,
the proposed adaptive preprocessing method achieves a visual token
reduction of approximately 55-58\% across most samples. The token
reduction pattern is highly consistent, indicating that adaptive
resolution selection and content-aware cropping operate reliably across
different document layouts. Images processed at lower resolutions and
with cropped spatial regions generate significantly fewer visual tokens,
directly reducing computational cost during vision encoding.

\begin{figure}[htbp]
    \centering
\includegraphics[width=\linewidth]{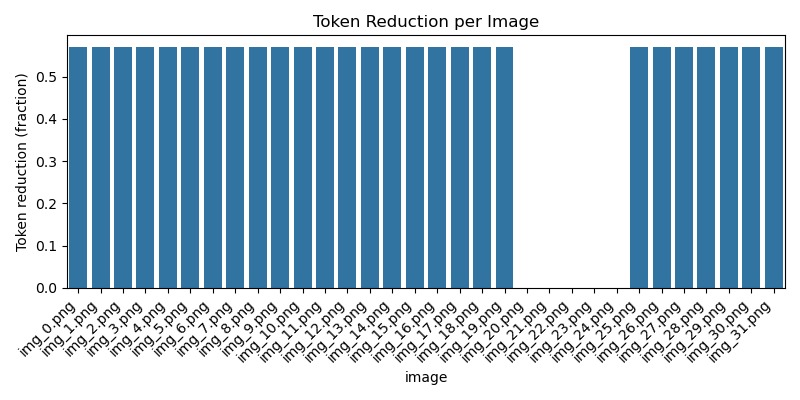}
    \caption{ Relative reduction in the number of visual tokens achieved by
the proposed adaptive preprocessing method compared to the baseline.}
    \label{fig:image9}
\end{figure}

For images identified as high complexity, visual token counts remain
closer to the baseline. This selective behavior ensures that the method
avoids excessive compression for images that require fine-grained visual
detail, while still providing substantial efficiency gains for simpler
inputs. Importantly, the magnitude of visual token reduction closely
aligns with the observed inference time improvements. Images with larger
token reductions consistently exhibit lower inference times, confirming
that visual token count is a primary driver of computational cost in the
FastVLM pipeline.

\emph{D. Relationship Between Visual Tokens and Visual Quality}

Fig. 8 illustrates the relationship between the number of visual tokens
generated after adaptive preprocessing and the corresponding visual
quality scores. Each data point represents one evaluated image processed
using the proposed method. The results show that a substantial reduction
in visual tokens, typically to fewer than 500 tokens per image, can be
achieved while maintaining high visual quality scores, with most samples
exhibiting quality values above 0.90. This indicates that adaptive
preprocessing effectively removes visually redundant information without
significantly degrading perceptual quality.

Specifically, the majority of evaluated images (over two-thirds of the
samples) maintain quality scores in the range of approximately
0.90--0.97, despite large variations in visual token counts. This
observation suggests that aggressive token reduction does not inherently
lead to quality degradation when guided by content-aware decisions. A
small number of samples exhibit lower quality scores, with values
dropping to approximately 0.70 or below. These cases correspond to
images with higher visual token counts (around 1,000 tokens), which are
identified as visually complex and therefore processed at higher
resolutions. This behavior aligns with the design of the proposed
method, which prioritizes preserving visual fidelity for complex
document images rather than enforcing uniform compression.

\begin{figure}[htbp]
    \centering
\includegraphics[width=\linewidth]{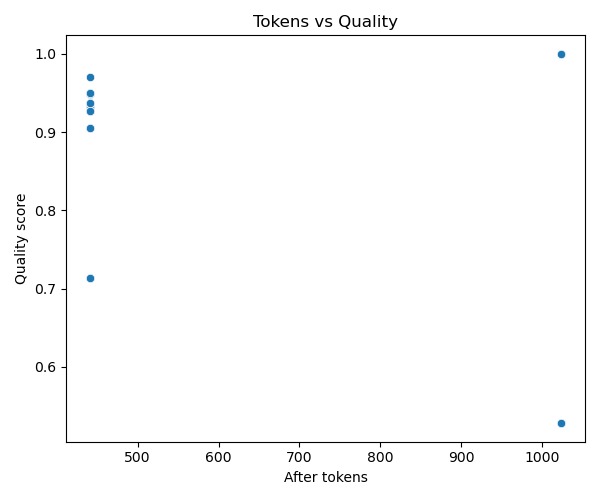}
    \caption{ Relationship between the number of visual tokens generated after
adaptive preprocessing and the corresponding visual quality scores,
showing that substantial token reduction can be achieved while
maintaining high visual quality for most samples.}
    \label{fig:image9}
\end{figure}

Overall, the scatter distribution shows no strong negative correlation
between visual token count and quality score. Instead, the results
suggest that visual quality remains largely stable across a wide range
of token reductions, supporting the claim that the proposed adaptive
preprocessing method improves efficiency without introducing severe
visual degradation required for downstream vision-language reasoning.

\emph{E. Qualitative Impact on VLM Output Stability}

Although this work primarily focuses on inference efficiency rather than
task-level accuracy, it is important to examine whether adaptive
preprocessing negatively impacts the semantic stability of VLM outputs.
In addition to qualitative inspection of generated answers, we analyze
per-image visual quality scores after adaptive preprocessing as a
supporting indicator of visual fidelity.

Fig. 9 reports the visual quality score for each evaluated image,
measured using structural similarity (SSIM) or a fallback quality metric
when SSIM is not applicable. The results show that the majority of
images retain high visual quality, with scores typically above 0.90,
despite substantial reductions in visual token count. This indicates
that adaptive preprocessing effectively removes visually redundant
regions while preserving content that is perceptually and semantically
meaningful.

\begin{figure}[htbp]
    \centering
\includegraphics[width=\linewidth]{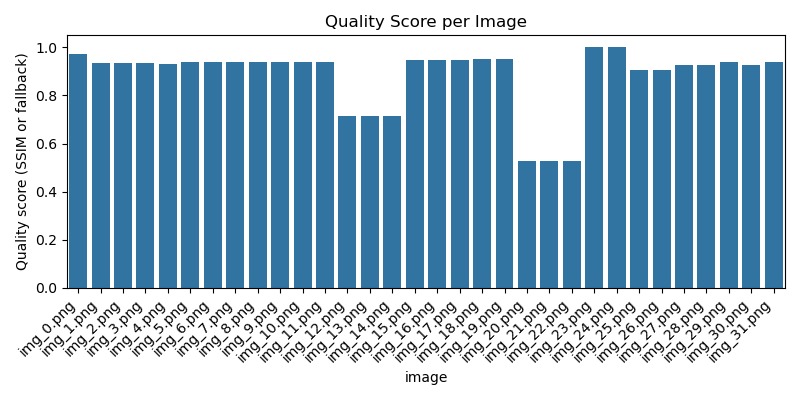}
    \caption{Per-image visual quality scores after adaptive preprocessing,
measured using structural similarity (SSIM) or fallback quality metrics.}
    \label{fig:image9}
\end{figure}

Notably, only a small subset of images exhibits lower quality scores,
which correspond to inputs with higher visual complexity. These images
are intentionally processed with higher resolutions under the adaptive
strategy, reflecting a conservative design choice that prioritizes
visual fidelity over aggressive compression. This behavior aligns with
the goal of maintaining stable vision-language reasoning rather than
maximizing token reduction uniformly.

To assess the impact on downstream reasoning, we further performed
qualitative inspection of generated VQA outputs under both baseline and
adaptive preprocessing pipelines. Across the evaluated samples, the
model produced semantically consistent answers, with no observable
degradation in reasoning correctness or answer relevance. Minor
variations were limited to surface-level phrasing differences rather
than semantic errors.

Overall, the combination of high visual quality scores and stable
generated answers suggests that the proposed adaptive preprocessing
improves inference efficiency without disrupting the core
vision-language alignment of FastVLM. These findings support the claim
that significant efficiency gains can be achieved by removing visual
redundancy at the input level while preserving information critical for
multimodal reasoning.
\end{quote}

\section{Discussion}
\begin{quote}
The experimental results indicate that adaptive visual preprocessing is
an effective strategy for improving inference efficiency in Fast
Vision-Language Models. Despite the architectural optimizations provided
by FastVLM, static preprocessing still incurs unnecessary visual
computation for many document images. By adapting input resolution and
spatial coverage based on image content, the proposed method
significantly reduces redundant visual processing without modifying the
underlying model.

A key observation from the results is the strong relationship between
visual token reduction and inference latency. The consistent decrease in
visual token count achieved by adaptive preprocessing closely aligns
with the observed reduction in inference time, confirming that visual
tokens are a major contributor to computational cost in the FastVLM
pipeline. By constraining token generation at the input level, the
proposed method effectively alleviates downstream computational overhead
during both vision encoding and language model prefilling.

Importantly, efficiency gains are achieved in a selective manner rather
than uniformly across all inputs. Images identified as visually complex
are intentionally processed at higher resolutions, resulting in
inference times closer to the baseline. This behavior reflects a
deliberate design choice that prioritizes robustness over aggressive
compression, ensuring that visually dense document images retain
sufficient detail for reliable vision-language reasoning.

The analysis of visual tokens and quality further supports this
selective strategy. The absence of a strong negative relationship
between token count and visual quality suggests that adaptive
preprocessing primarily removes redundant visual information rather than
degrading semantically meaningful content. Together, these findings
demonstrate that input-aware preprocessing provides a practical and
lightweight complement to architectural efficiency improvements,
offering a balanced trade-off between computational efficiency and
visual fidelity. These findings suggest that input-aware preprocessing
represents a largely underexplored but highly impactful axis for
efficiency optimization in vision-language systems.

More broadly, the results highlight a complementary perspective on
efficiency optimization in vision-language models. While existing work
predominantly emphasizes architectural or model-level solutions, this
study shows that substantial efficiency gains can be achieved through
preprocessing alone. By operating entirely at the input level, the
proposed method avoids retraining or architectural modification, making
it particularly suitable for practical deployment. These findings
suggest that preprocessing-level adaptation remains an underexplored yet
impactful direction for improving efficiency in vision-language systems.

From a reasoning perspective, the stability of generated answers
indicates that adaptive preprocessing does not interfere with the core
vision-language alignment learned by FastVLM. Since preprocessing
operates solely on visual redundancy rather than semantic content, the
language model continues to receive sufficient visual cues for effective
reasoning. This supports the hypothesis that dense visual tokens are
often unnecessary for correct multimodal inference, particularly in
document-centric tasks.

The proposed adaptive preprocessing framework is broadly applicable to
vision-language models that employ patch-based or tokenized visual
encoders. Models such as LLaVA and BLIP-2, which rely on
fixed-resolution visual embeddings, can directly benefit from reduced
visual token counts through input-level adaptation. However, the method
may be less effective for architectures that internally perform dynamic
visual token selection or rely on end-to-end learned visual compression.
Additionally, task domains requiring fine-grained pixel-level precision,
such as small-object detection, may require more conservative threshold
settings. These limitations highlight opportunities for future work that
jointly considers preprocessing-level adaptation and model-level
sparsification.
\end{quote}

\section{Conclusion}
This paper presented an adaptive visual preprocessing method to improve
the inference efficiency of Fast Vision-Language Models. By dynamically
selecting input resolution and applying content-aware cropping, the
proposed approach reduces visual redundancy prior to vision encoding
without modifying the FastVLM architecture or requiring retraining.
Experimental results show substantial efficiency gains, including over
50\% reduction in per-image inference latency, lower mean full
generation time, and a significant decrease in visual token usage
compared to the baseline pipeline. These improvements are achieved
through selective computation that preserves higher-resolution
processing for visually complex inputs while avoiding unnecessary
computation for simpler cases. Overall, this work demonstrates that
input-aware preprocessing constitutes an effective and practical
complement to architectural efficiency improvements. Future work may
investigate the integration of adaptive preprocessing with model-level
sparsification techniques to further enhance deployment-oriented
performance. Overall, this study demonstrates that substantial inference
efficiency gains can be achieved without architectural modification or
retraining, reinforcing input-aware preprocessing as a practical and
generalizable strategy for deployment-oriented vision-language systems.

\section*{References}

\begin{enumerate}
\def\labelenumi{\arabic{enumi}.}
\item
  \begin{quote}
  T. Brown \emph{et al.}, ``Language Models are Few-Shot Learners,''
  \emph{Advances in Neural Information Processing Systems (NeurIPS)},
  2020.
  \end{quote}
\item
  \begin{quote}
  H. Touvron \emph{et al.}, ``LLaMA: Open and Efficient Foundation
  Language Models,'' \emph{arXiv preprint arXiv:2302.13971}, 2023.
  \end{quote}
\item
  \begin{quote}
  J. Li \emph{et al.}, ``BLIP-2: Bootstrapping Language-Image
  Pre-training with Frozen Image Encoders and Large Language Models,''
  \emph{Proceedings of the IEEE/CVF Conference on Computer Vision and
  Pattern Recognition (CVPR)}, 2023.
  \end{quote}
\item
  \begin{quote}
  A. Radford \emph{et al.}, ``Learning Transferable Visual Models from
  Natural Language Supervision,'' \emph{Proceedings of the International
  Conference on Machine Learning (ICML)}, 2021.
  \end{quote}
\item
  \begin{quote}
  J.-B. Alayrac \emph{et al.}, ``Flamingo: a Visual Language Model for
  Few-Shot Learning,'' \emph{Proceedings of the Advances in Neural
  Information Processing Systems (NeurIPS)}, 2022.
  \end{quote}
\item
  \begin{quote}
  J. Kaplan \emph{et al.}, ``Scaling Laws for Neural Language Models,''
  \emph{arXiv preprint arXiv:2001.08361}, 2020.
  \end{quote}
\item
  \begin{quote}
  D. Driess \emph{et al.}, ``PaLM-E: An Embodied Multimodal Language
  Model,'' \emph{arXiv preprint arXiv:2303.03378}, 2023.
  \end{quote}
\item
  \begin{quote}
  P. K. A. Vasu \emph{et al.}, ``FastVLM: Efficient Vision Encoding for
  Vision-Language Models,'' \emph{Proceedings of the IEEE/CVF Conference
  on Computer Vision and Pattern Recognition (CVPR)}, 2025.
  \end{quote}
\item
  \begin{quote}
  M. Rang \emph{et al.}, ``Eve: Efficient Multimodal Vision-Language
  Models with Elastic Visual Experts,'' \emph{arXiv preprint}, 2025.
  \end{quote}
\item
  \begin{quote}
  H. Liu \emph{et al.}, ``Visual instruction tuning,'' \emph{arXiv
  preprint} arXiv:2304.08485, 2023.
  \end{quote}
\item
  \begin{quote}
  H. Liu \emph{et al.}, ``Improved baselines with visual instruction
  tuning,'' \emph{arXiv preprint} arXiv:2310.03744, 2023.
  \end{quote}
\item
  \begin{quote}
  W. Dai \emph{et al.}, ``InstructBLIP: Towards general-purpose
  vision-language models with instruction tuning,'' in \emph{Advances in
  Neural Information Processing Systems (NeurIPS)}, 2023.
  \end{quote}
\item
  \begin{quote}
  X. Chu \emph{et al.}, ``MobileVLM V2: Faster and stronger baseline for
  vision-language models,'' \emph{arXiv preprint} arXiv:2402.03766,
  2024.
  \end{quote}
\item
  \begin{quote}
  C. Ge \emph{et al.}, ``ConvLLaVA: Hierarchical backbones as visual
  encoder for large multimodal models,'' \emph{arXiv preprint}
  arXiv:2405.15738, 2024.
  \end{quote}
\item
  \begin{quote}
  J. Chen \emph{et al.}, ``ViTamin: Designing scalable vision models in
  the vision-language era,'' \emph{arXiv preprint} arXiv:2404.02132,
  2024.
  \end{quote}
\item
  \begin{quote}
  Y. Shang \emph{et al.}, ``LLaVA-PruMerge: Adaptive token reduction for
  efficient large multimodal models,'' \emph{arXiv preprint}
  arXiv:2403.15388, 2024.
  \end{quote}
\item
  \begin{quote}
  L. Chen \emph{et al.}, ``An image is worth 1/2 tokens after layer 2:
  Plug-and-play inference cost optimization for large vision-language
  models,'' in \emph{European Conference on Computer Vision (ECCV)},
  2024.
  \end{quote}
\item
  \begin{quote}
  Y. Zhang \emph{et al.}, ``SparseVLM: Visual token sparsification for
  efficient vision-language model inference,'' \emph{arXiv preprint}
  arXiv:2410.04417, 2024.
  \end{quote}
\item
  \begin{quote}
  Z. Guo \emph{et al.}, ``LLaVA-UHD: An LMM perceiving any aspect ratio
  and high-resolution images,'' in \emph{European Conference on Computer
  Vision (ECCV)}, 2024.
  \end{quote}
\item
  \begin{quote}
  W. Huang \emph{et al.}, ``Dynamic-LLaVA: Efficient multimodal large
  language models via dynamic vision-language context sparsification,''
  \emph{arXiv preprint} arXiv:2412.00876, 2024.
  \end{quote}
\item
  \begin{quote}
  M. Mathew et al., ``DocVQA Dataset,'' available:
  https://www.docvqa.org/datasets
  \end{quote}
\end{enumerate}

\end{document}